\documentclass[9pt,conference]{IEEEtran}
\IEEEoverridecommandlockouts
\usepackage{cite}
\usepackage{amsmath,amssymb,amsfonts}
\usepackage{algorithmic}
\usepackage{graphicx}
\usepackage{textcomp}
\usepackage{xcolor}
\usepackage{amssymb}
\usepackage{graphicx}

\def\BibTeX{{\rm B\kern-.05em{\sc i\kern-.025em b}\kern-.08em
    T\kern-.1667em\lower.7ex\hbox{E}\kern-.125emX}}
\begin{document}

\title{Seeing Beyond Frames: Zero-Shot Pedestrian Intention Prediction with Raw Temporal Video and Multimodal Cues
{\footnotesize \textsuperscript{*} }
\thanks{ }
}

\author{\IEEEauthorblockN{1\textsuperscript{st}Pallavi Zambare}
\IEEEauthorblockA{\textit{Departmrnt of computer science} \\
\textit{Texas Tech University}\\
Lubbock, USA \\
pzambare@ttu.edu}
\and
\IEEEauthorblockN{2\textsuperscript{nd}Venkata Nikhil Thanikella}
\IEEEauthorblockA{\textit{Departmrnt of computer science} \\
\textit{Texas Tech University}\\
Lubbock, USA \\
vthanike@ttu.edu}
\and
\IEEEauthorblockN{3\textsuperscript{rd} Ying Liu }
\IEEEauthorblockA{\textit{Departmrnt of computer science} \\
\textit{Texas Tech University}\\
Lubbock, USA \\
Y.Liu@ttu.edu}
\and
}
\maketitle

\begin{abstract}
Pedestrian intention prediction is essential for autonomous driving in complex urban environments. Conventional approaches depend on supervised learning over frame sequences and require extensive retraining to adapt to new scenarios. Here, we introduce BF PIP (Beyond Frames Pedestrian Intention Prediction), a zero-shot approach built upon Gemini 2.5 Pro. It infers crossing intentions directly from short, continuous video clips enriched with structured JAAD metadata. In contrast to GPT-4V–based methods that operate on discrete frames, BF-PIP processes uninterrupted temporal clips. It also incorporates bounding-box annotations and ego-vehicle speed via specialized multimodal prompts. Without any additional training, BF-PIP achieves 73\% prediction accuracy, outperforming a GPT-4V baseline by 18 \%. These findings illustrate that combining temporal video inputs with contextual cues enhances spatiotemporal perception and improves intent inference under ambiguous conditions. This approach paves the way for agile, retraining-free perception module in intelligent transportation system.
 \end{abstract}


\section{Introduction}
As autonomous vehicles (AVs) increasingly become a reality in modern transportation. Consequently, the necessity for an accurate and timely understanding of pedestrian behavior has emerged as a critical safety requirement. Among the various behavioral cues, pedestrian crossing intention prediction is mainly essential for anticipating potential road conflicts and ensuring safe navigation in urban environments \cite{b1,b2}. Early research relied on handcrafted features such as trajectories, body orientation, and gaze, processed with RNNs or LSTMs to capture temporal dynamics \cite{b3, b4} Recently, approaches like Pedestrian Graph+ and ST\-CrossingPose employed Graph Convolutional Networks (GCNs) to represent spatial relationships and enhance pose\-based inference \cite{b5,b6}. Transformer based methods like PIT-Block and IntFormer introduced enhanced spatiotemporal attention and outperformed earlier sequence models\cite{b7, b8}. 

Despite these advances, supervised methods depend on static image sequences, require extensive labeled data, and generalize poorly to novel environments  \cite{b2, b9, b10}. To overcome these limitations, Multimodal Large Language Models (MLLMs) such as GPT-4V, LLAVA, and GPT-4o have emerged. They offer robust zero-shot reasoning across vision and language inputs \cite{b11, b12, b13}. OmniPredict  \cite{b14}, a new benchmark using GPT-4o, established the probability of instruction-prompted multimodal inference. It predicted pedestrian behavior using scene images, bounding boxes, and vehicle speed. Although effective, it still processes discrete frame sequences, restraining its ability to capture motion cues like hesitation, body shifts, or gaze changes \cite{b15, b16}.

Even at its best, OmniPredict employs static frame sequences and treats the scene as a series of discrete observations. This limits the temporal continuity that is naturally captured in video. In contrast, video-based inputs can capture motion dynamics, hesitation, gaze shifts, and interaction cues that are not easily inferred from still frames.
\begin{figure}[htbp]
\centerline{\includegraphics[width=8.5cm,height=5 cm]{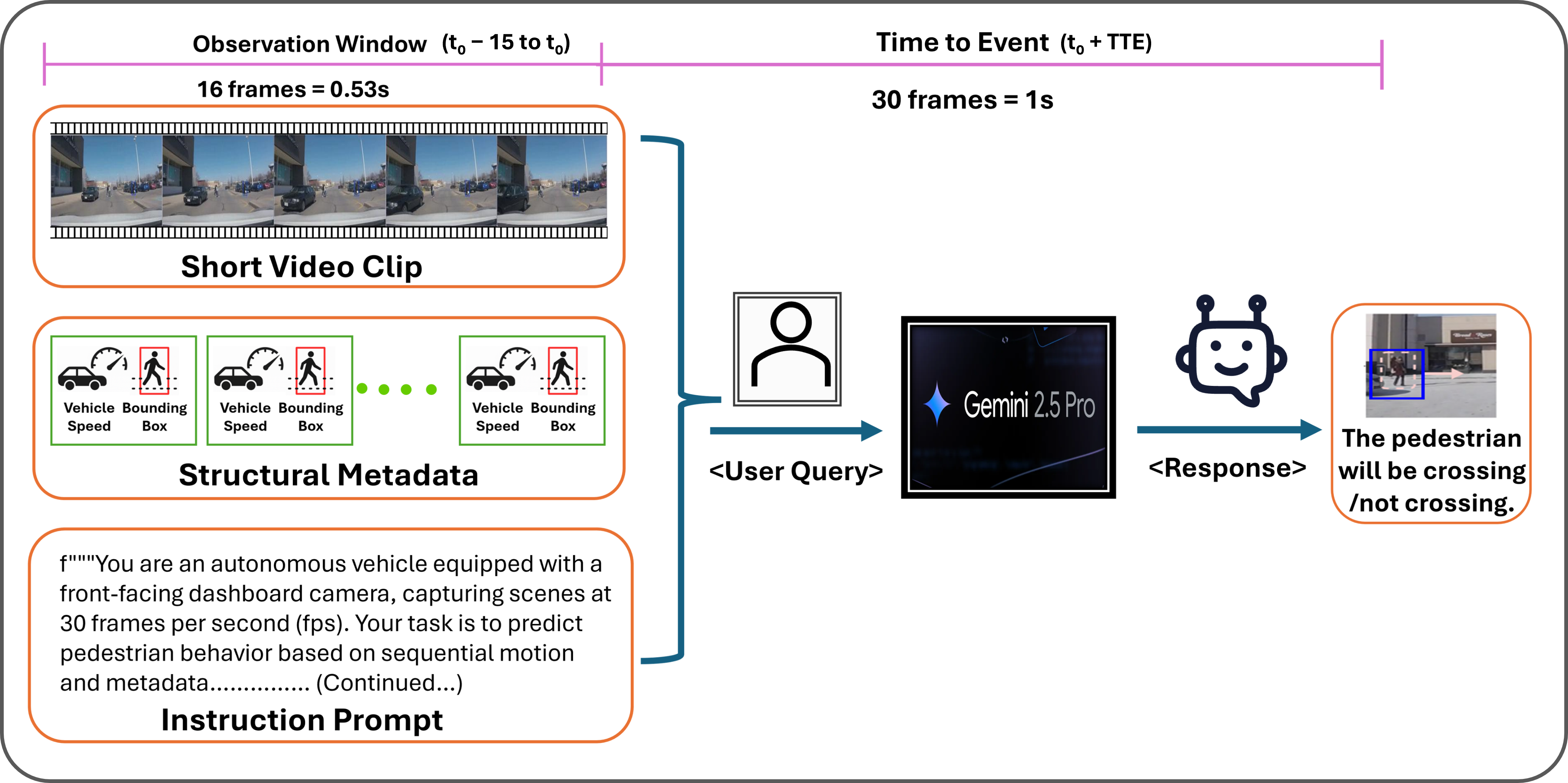}}
\caption{BF-PIP Framework}
\label{fig}
\end{figure}

In this work, we introduce BF-PIP, a zero-shot pedestrian intention prediction framework that leverages short video clips and along with structured metadata (bounding boxes and ego-vehicle speed) in a multimodal prompt to Gemini 2.5 Pro. In contrast to previous methods that relied on image sequences, our method allows for continuous motion capture, enabling spatiotemporal reasoning without the need for retraining.  Evaluation was conducted using the JAAD$\textsubscript{beh}$ \cite{b19} dataset, a widely recognized benchmark in autonomous driving research. BF-PIP reached an impressive 73\% accuracy in a zero-shot setting, representing an 18\% improvement over the performance of GPT4V-PBP \cite{b4} and outperforming OmniPredict\cite{b14} by 6\%, which is currently the leading MLLM-based approach. Without requiring additional training, BF-PIP outperforms domain-specific models in predicting pedestrian intent. Qualitative analysis confirms Gemini 2.5 Pro’s ability to interpret complex scenes using spatial and behavioral cues. An ablation study further quantifies the contribution of each input modality.

In conclusion, BF-PIP marks a significant advance beyond traditional vision-based models by directly analyzing continuous video streams with structured metadata. It operates within a prompt-driven, zero-shot framework. Results indicate that BF-PIP’s ability to process raw video input reduces the need for extensive preprocessing. This capability enables zero-shot intent prediction in unfamiliar environments, contributing to more efficient and safer autonomous driving operations.

\section{Methodology}
Gemini 2.5 Pro represents a significant advancement in multimodal AI, offering the ability to process and reason over rich combinations of video, image, and text data through a single prompt interface. Unlike previous models, Gemini 2.5 Pro natively supports raw video input, enabling temporally grounded understanding of pedestrian motion and scene context. In our experiments, we used the Google Gemini API to perform zero-shot pedestrian intention prediction tasks on the JAAD dataset \cite{b2}through the Gemini 2.5 Pro model. Specifically, we focused on the JAADbeh subset \cite{b2}, which includes detailed annotations of pedestrian behavior and environmental factors.

\subsection{Task Identification}
Pedestrian crossing intention prediction is formulated as a binary classification task: the model must determine whether a pedestrian will cross or not cross the road within a fixed future time horizon. Given a temporal sequence of observations captured from the ego-vehicle’s front-facing camera, the goal is to anticipate the pedestrian’s crossing action 30 frames (1 second) into the future, referred to as the Time-To-Event (TTE). For each pedestrian instance, an observation window of 16 frames (approximately 0.5 seconds) immediately preceding the TTE point is defined. These frames capture the pedestrian’s motion history and environmental context leading up to the moment of prediction. This temporal segment is extracted following the JAAD benchmark protocol \cite{b17,b2}, where t$_0$ represents the last observed frame and prediction occurs at t$_0$+ TTE.

The task is defined under two input configurations: annotated and unannotated. In the annotated setting, short video clips are provided with per-frame pedestrian bounding boxes from the JAAD dataset, preserving temporal continuity and enabling precise spatial localization. In contrast, the unannotated setting involves raw video clips without bounding boxes, allowing assessment of Gemini 2.5 Pro’s ability to infer pedestrian intent solely from visual motion and contextual cues.
\subsection{Input Modalities}
The BF-PIP pipeline accepts three main types of input data to support pedestrian intention prediction. These modalities are incorporated either as visual embeddings within the video stream or as structured metadata attached to each instance.
\begin{itemize}
    \item \textbf{Short Video Clip}: A temporally continuous short clip is constructed from the JAAD video sequences, capturing the target pedestrian within a 16-frame observation window. These clips maintain motion continuity and more closely resemble real-world perception compared to static image sequences.
    \item \textbf{Bounding Box Coordinates}: Each frame includes bounding box annotations (x, y, width, height) extracted from JAAD metadata. These are either rendered into the clip (for the annotated case) or provided as separate input (for the structured mode). They help localize the pedestrian in the scene and focus the model’s attention.
     \item \textbf{Vehicle Speed}: At each frame, vehicle speed is included as part of the ego-motion metadata to support contextual reasoning. For JAAD, categorical ego-speed values are used to indicate the trend of the vehicle's motion (e.g., decelerating, maintaining constant speed, or accelerating). 
\end{itemize}
This multimodal input (video, spatial annotations, and motion data) is embedded into a prompt that's given to Gemini 2.5 Pro. The model is instructed to reason over these inputs and make a binary decision.
\subsection{Prompt Construction and Inference Strategy}
 To enable zero-shot reasoning, a structured prompt is designed to embed the task context and input modality definitions, allowing Gemini 2.5 Pro to perform multimodal understanding directly from short video clips and associated metadata. The prompt consists of two stages:\\
\textbf{ Stage 1: Environment and Task Setup}: \\
The first part of the prompt presents Gemini 2.5 Pro to its role as an autonomous vehicle equipped with a front-facing camera. It specifies the temporal nature of the input (a 16-frame video segment recorded at 30 FPS). It describes the metadata included per frame, including bounding boxes and ego-vehicle motion. The task is presented as a binary classification problem, requiring a decision on whether the observed pedestrian will cross the road.
\begin{figure}[htbp]
\centerline{\includegraphics[width=8.5cm,height=4cm]{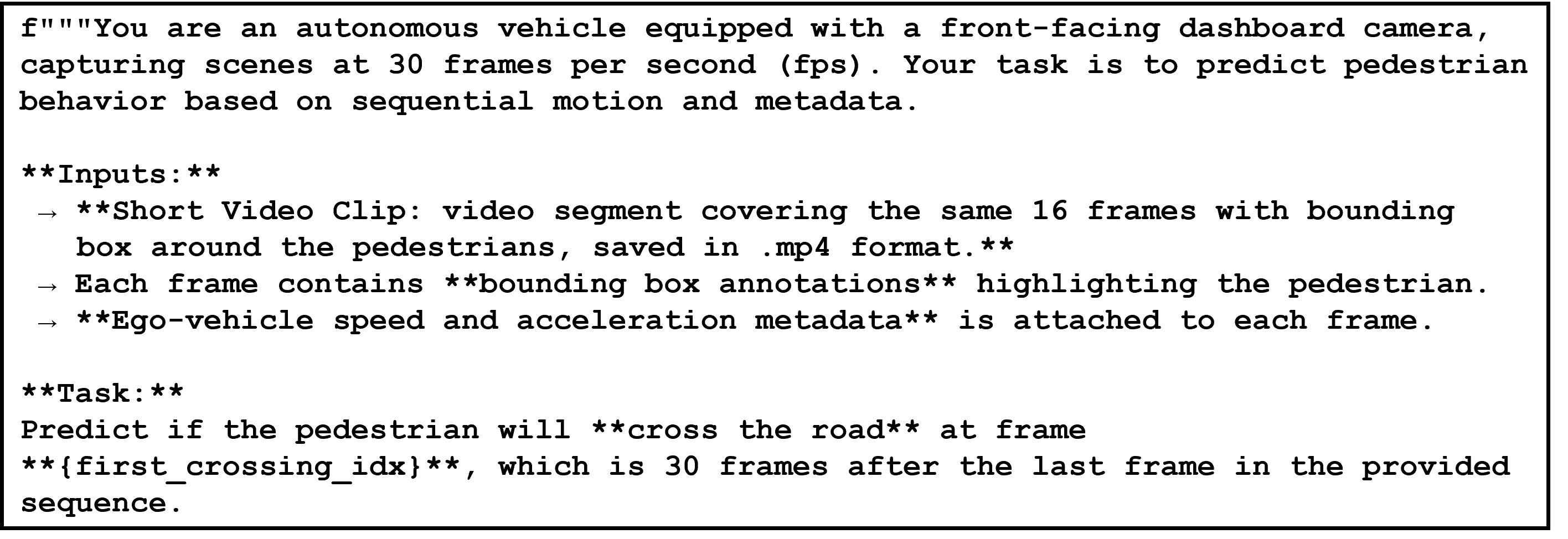}}
\label{fig}
\end{figure}
 
\textbf{Stage 2: Behavioral Reasoning and Output Constraints}:\\
The second half of the prompt incorporates explicit reasoning steps to guide internal decision-making. These steps include the analysis of pedestrian posture, movement patterns, and surrounding visual cues. The classification labels are explicitly defined for the model. To ensure consistency and improve interpretability, the output is constrained to a single-word prediction.
\begin{figure}[htbp]
\centerline{\includegraphics[width=8.5cm,height=6cm]{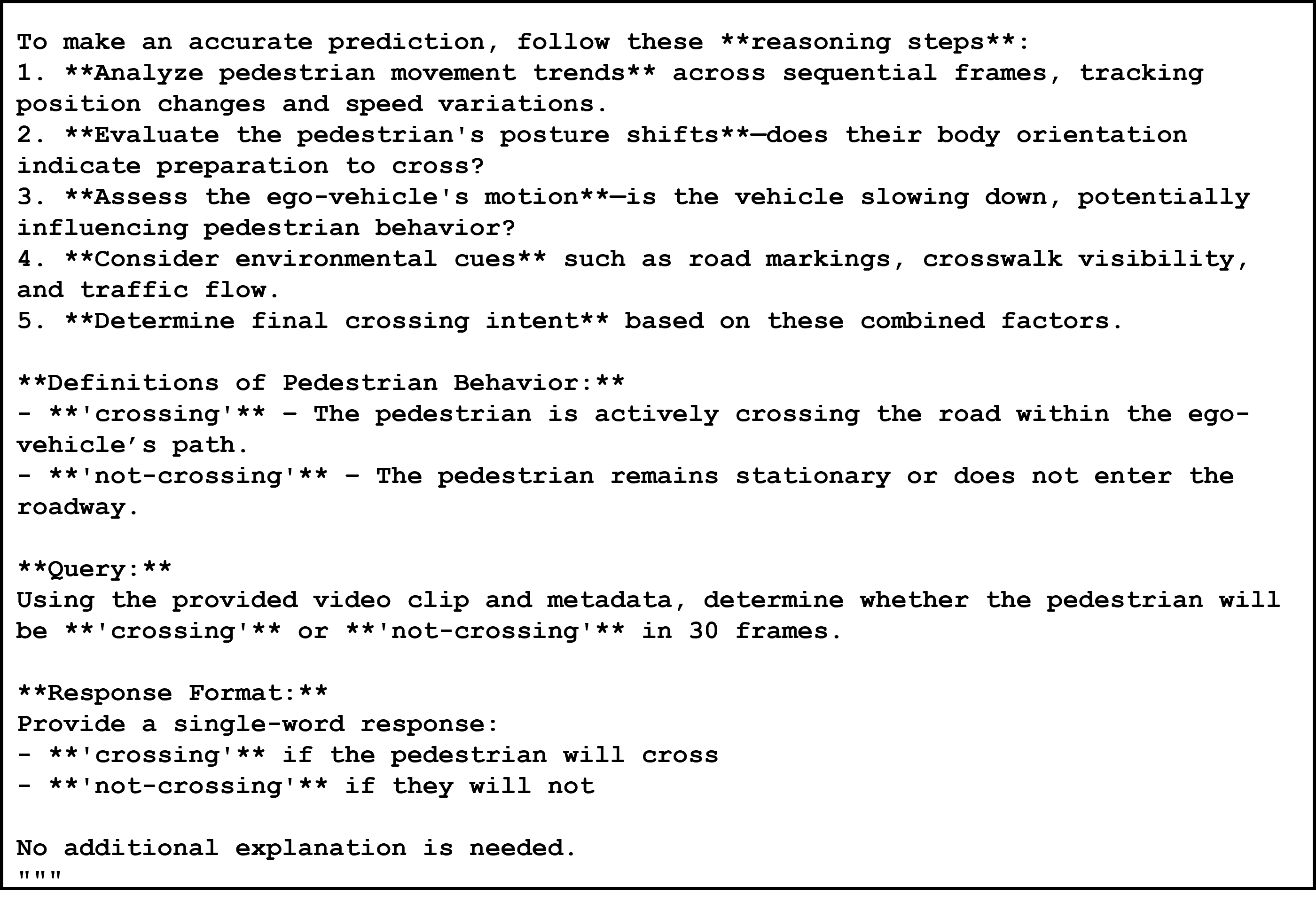}}
\label{fig}
\end{figure}
This structure for prompts was uniformly applied across all evaluation conditions, making them consistent and reproducible. The prompting logic that was built in enables Gemini 2.5 Pro to use its multimodal attention over visual and textual inputs to infer pedestrian intention without the use of fine-tuning. A strategy employing role-play prompting \cite{b19} is used to enhance interpretability and decision-making. The role that Gemini 2.5 Pro plays is that of a situational observer, systematically analyzing such things as pedestrian motion cues, environmental conditions, and past behavioral patterns. By framing its role explicitly as an “intention predictor,” the model generates responses with more contextual awareness.

This role-play prompting strategy serves two primary objectives. First, it encourages behavioral alignment by reasoning rooted in human-like observational strategies, enabling predictions to align with real-world pedestrian behavior. Second, it enhances Contextual Fusion by multimodal integration by reinforcing attention on bounding box dynamics, trajectory shifts, and implicit social cues. Additionally, this approach complements chain-of thought prompting \cite{b20}, allowing the model to articulate stepwise decision-making before arriving at a binary prediction. By integrating role-play with structured multimodal inputs, reasoning depth and reliability are improved in zero-shot settings

By leveraging Gemini 2.5 Pro’s zero-shot multimodal reasoning capability, our method avoids task-specific training. Instead, it relies on carefully constructed prompts to guide pedestrian intention prediction. This prompt-based framework accepts both annotated and unannotated short video clips as input. It supports robust decision-making grounded in a temporally rich context. The integration of role-play framing, multimodal inputs, and structured reasoning yields a generalizable pipeline capable of handling diverse street scenes, making it well-suited for real-world autonomous driving scenarios.

In the last step of the procedure, Gemini 2.5 Pro is instructed to produce predictions in a structured JSON format to ensure consistency in output interpretation and evaluation. since Gemini 2.5 Pro is a generative model, whose outputs may vary across runs. To enhance determinism, the model is configured with a temperature of 0 and a fixed random seed of 0. Each prompt is executed five times per instance to evaluate response stability and ensure reliability.
\renewcommand{\arraystretch}{1.2}  
\label{tab:model_comparison}
\begin{table*}[!htbp]
\caption{Performance comparison with state-of-the-art methods from OmniPredict \cite{}. The input modalities represent different data types: I: image, B: bounding box coordinates, P: skeleton-based pose, S: vehicle speed, and V: video input (short temporal clip).}
\centering
\resizebox{\textwidth}{!}{%
\begin{tabular}{|l|c|l|*{6}{c}|*{5}{c}|}
\hline
\textbf{Models} & \textbf{Year} & \textbf{Model Variants} & \multicolumn{6}{c|}{\textbf{Inputs}} & \multicolumn{5}{c|}{\textbf{JAAD-beh}} \\
\cline{4-9} \cline{10-14}
 & & & \textbf{I} & \textbf{B} & \textbf{P} & \textbf{S} & \textbf{V} & \textbf{Extra Info.} & \textbf{ACC} & \textbf{AUC} & \textbf{F1} & \textbf{P} & \textbf{R} \\
\hline
MultiRNN \cite{b3} & 2018 & GRU & \checkmark & \checkmark & \checkmark & \checkmark & -- & -- & 0.61 & 0.50 & 0.74 & 0.64 & 0.86 \\
SFRNN \cite{b4} & 2020 & GRU & \checkmark & \checkmark & \checkmark & \checkmark & -- & -- & 0.51 & 0.45 & 0.63 & 0.61 & 0.64 \\
SingleRNN \cite{b2} & 2020 & GRU & \checkmark & \checkmark & \checkmark & \checkmark & -- & -- & 0.58 & 0.54 & 0.67 & 0.67 & 0.68 \\
PCPA \cite{b17}& 2021 & RNN+Attention & \checkmark & \checkmark & \checkmark & \checkmark & -- & -- & 0.58 & 0.50 & 0.71 & -- & -- \\
IntFormer \cite{b8} & 2022 & Transformer & \checkmark & \checkmark & \checkmark & \checkmark & -- & -- & 0.59 & 0.54 & 0.69 & -- & -- \\
ST CrossingPose \cite{b6} & 2022 & Graph CNN & \checkmark & \checkmark & \checkmark & -- & -- & -- & 0.63 & 0.56 & 0.74 & 0.66 & 0.83 \\
FFSTP \cite{b16} & 2022 & GRU+Attention & \checkmark & \checkmark & \checkmark & \checkmark & -- & -- & 0.62 & 0.54 & 0.74 & 0.65 & 0.85 \\
Pedestrian Graph+ \cite{b5} & 2022 & Graph CNN+Attention & \checkmark & \checkmark & \checkmark & \checkmark & -- & -- & 0.70 & 0.70 & 0.76 & 0.77 & 0.75 \\
PIT-Block(a) \cite{b7} & 2022 & Transformer & \checkmark & \checkmark & \checkmark & \checkmark & -- & -- & 0.70 & 0.65 & 0.81 & 0.71 & 0.93 \\
GPT4V-PBP \cite{b15} & 2023 & MLLM & \checkmark & \checkmark & -- & -- & -- & Text & 0.57 & 0.61 & 0.65 & 0.82 & 0.54 \\
GPT4V-PBP Skip \cite{b15}& 2023 & MLLM & \checkmark & \checkmark & -- & -- & -- & Text & 0.55 & 0.59 & 0.64 & 0.81 & 0.53 \\
OmniPredict \cite{b14} & 2024 & MLLM & \checkmark & \checkmark & -- & \checkmark & -- & Text & 0.67 & 0.65 & 0.65 & 0.66 & 0.65 \\
BF-PIP(Ours) & 2025 & MLLM & -- & \checkmark & -- & \checkmark & \checkmark & Text & 0.73 & 0.77 & 0.80 & 0.96 & 0.69 \\
\hline
\end{tabular}%
}
\label{tab:model_comparison}
\end{table*}

\section{Experiment and results}
\subsection{Dataset}
The experiments were conducted on the JAAD$\textsubscript{beh}$ subset of the Joint Attention in Autonomous Driving (JAAD) dataset \cite{b2}. It was recorded using front-facing dashboard cameras mounted on vehicles in diverse urban environments. The dataset was specifically organized to assess pedestrian intention in realistic traffic scenarios. The broader JAADall dataset includes all pedestrian instances, irrespective of crossing behavior. In contrast, JAAD$\textsubscript{beh}$ was a focused subset containing 686 pedestrian instances who were either actively crossing or exhibiting intent to cross, each accompanied by detailed behavioral annotations.The full JAAD dataset contains 346 pedestrian video clips, divided into 188 for training, 32 for validation, and 126 for testing. For evaluation, only the 126 test clips were utilized. These test samples align with established benchmark configurations and were used for zero-shot inference. They also enable direct comparison with existing models such as PCPA and OmniPredict.

For each annotated pedestrian, a short video segment composed of 16 frames was extracted, ending 30 frames before the predicted Time-To-Event (TTE). The evaluation was conducted in two modes.
\begin{itemize}
    \item \textbf{Annotated mode}: Bounding boxes are rendered on each frame to indicate the pedestrian of interest.
    \item \textbf{Unannotated mode}: The same clips are provided without any visual overlays. This allows us to assess Gemini 2.5 Pro’s capacity to infer pedestrian intent purely from raw visual input.
\end{itemize}
This dual-mode setup enables comprehensive evaluation under both structured and perceptual reasoning conditions. It also facilitates fair comparisons with both vision-based and prompt-driven models.
\subsection{Evaluation metric}
To evaluate how effectively pedestrian crossing intention was predicted in comparison to benchmark models, five standard classification metrics were used: Accuracy (ACC), Precision (P), Recall (R), F1 Score (F1), and Area Under the ROC Curve (AUC). These metrics provide a comprehensive assessment of overall model performance. They also capture class-wise discrimination, ensuring a balanced evaluation across varied crossing behaviors and prediction outcomes.

\subsection{Implementation Setup}
The Gemini 2.5 Pro–based evaluation pipeline was deployed using services provided by Google Cloud Platform (GCP). A dedicated GCP project was initialized with Vertex AI and Cloud Storage APIs enabled. A service account was configured with appropriate IAM roles (Vertex AI User and Storage Object Admin) to enable access to the Gemini API and Cloud Storage. All short video clips were uploaded to a Cloud Storage bucket and retrieved dynamically during inference.

For local execution, the environment was configured using standard GCP authentication and project variables to enable access to Vertex AI and Cloud Storage services. During each evaluation instance, Gemini 2.5 Pro received a structured prompt along with the corresponding video segment through the GenAI API. The prediction output was generated under the deterministic setup described in the methodology. It was then recorded and used for metric computation.
\subsection{Quantitative results}
To quantitatively assess model performance, Table 1 presents a comparative evaluation of the proposed BF-PIP model. It is compared against both classical domain-specific baselines and recent MLLM-based approaches on the JAADbeh benchmark. All models were predicting pedestrians crossing 30 frames ahead using 16 past frames. However, the GPT4V-PBP variants rely on 10-frame inputs. Conversely, the proposed BF-PIP model operates directly on short, continuous video clips to capture richer temporal context. The table also includes each model’s release year and the input modalities used for prediction. Model performance is evaluated using the five metrics introduced earlier: Accuracy (ACC), Area Under the ROC Curve (AUC), F1 Score (F1), Precision (P), and Recall (R).

Even without additional training, Gemini-based MLLM models establish competitive performance. In particular, the proposed BF-PIP model achieves the highest accuracy 0.73 and an AUC of 0.76. It also attains a strong F1 score of 0.80, which is only 0.01 lower than benchmark model PIT-Block(a). Importantly, BF-PIP records the highest precision 0.96, emphasizing its reliability in positive predictions. Additionally, it maintains a recall of 0.68, comparable to or exceeding existing Transformer-based baselines. Remarkably, BF-PIP achieves these results while relying on significantly fewer specialized features than other methods. These results demonstrate that BF-PIP’s hybrid use of bounding-box-guided attention and short-clip visual context is highly effective. It yields a new state of the art in pedestrian crossing intention prediction. It effectively balances high overall accuracy, low uncertainty, and strong reliability in positive class identification.

\subsection{Qualitative results}
A qualitative analysis was conducted to examine how Gemini 2.5 Pro reasons interprets pedestrian behavior from video input. The study assessed the model’s attention to motion continuity, posture changes, and spatial context.

As illustrated in Fig. 2, the pedestrian in the red box was initially positioned on the sidewalk near a marked crosswalk. The model demonstrates strong contextual understanding by concurrently identifying relevant road users and analyzing traffic dynamics, such as scanning for oncoming vehicles. It also extracts environmental cues, including crosswalk markings and visual obstructions. Rather than treating all pedestrians equally, Gemini 2.5 Pro gives higher priority to individuals positioned closer to the roadway. Pedestrians standing near the curb edge receive more attention. The model also focuses on subtle behavioral cues such as leaning forward and looking toward traffic. Additionally, it considers small but decisive steps onto the crosswalk as strong indicators that the pedestrian is ready to cross. It also focuses on subtle behavioral cues, including posture shifts (a forward lean), gaze direction (observing traffic), and micro-movements (a decisive step onto the crosswalk) that indicate readiness to cross. By fusing spatial, temporal, and contextual information, the model generates robust predictions of pedestrian intention. This enables downstream planning modules to make more reliable and risk-aware decisions.
\begin{figure}[htbp]
\centerline{\includegraphics[width=8.5cm,height=5.5cm]{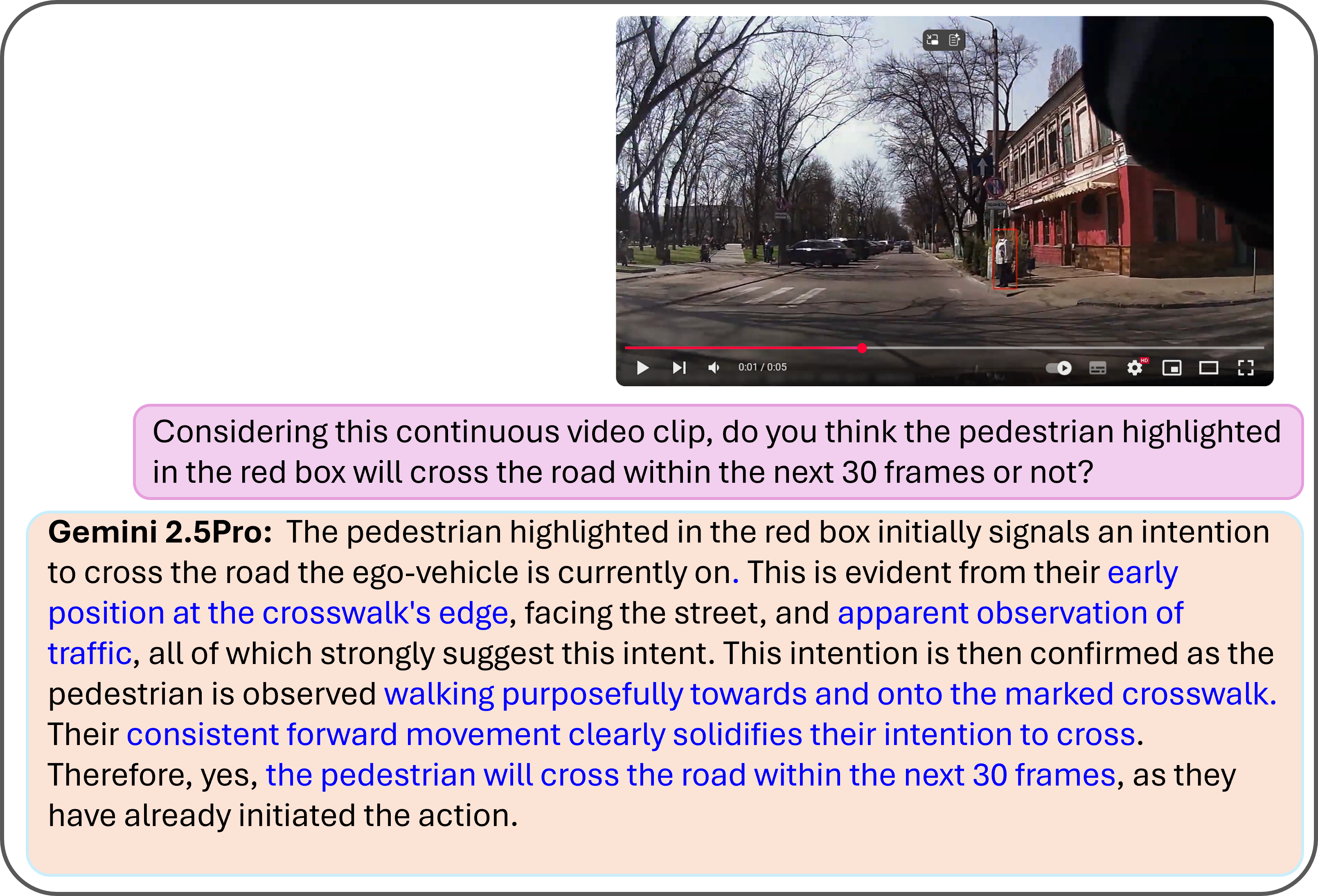}}
\caption{Pedestrian crossing intention prediction.}
\label{fig}
\end{figure}

\subsection{Ablation Study}
An ablation study was conducted to isolate the impact of each input modality on BF-PIP performance. The 8 configurations were evaluated, including unannotated video (UV), annotated video (AV), bounding-box coordinates (BB), and ego-vehicle speed (S), both individually and in all pairwise combinations. All model predictions concern the crossing intent 30 frames ahead under identical conditions. Results are measured using the five metrics mentioned previously and are summarized in Table II.
\begin{table}[!htpb]
\centering
\caption{Ablation study of input modalities. UV: unannotated video, AV: annotated video, BB: bounding box coordinates, S: ego-vehicle speed.}
\begin{tabular}{|l|c|c|c|c|c|}
\hline
\textbf{Input Modality} & \textbf{ACC} & \textbf{AUC} & \textbf{F1} & \textbf{P} & \textbf{R} \\
\hline
UV                      &      0.65 &  0.62      &  0.74  & 0.96  & 0.60   \\
UV + S                  &      0.70 &  0.74     &  0.78  & 0.97 & 0.65   \\
UV + BB                 &      0.60 & 0.58       &  0.68  & 0.96  & 0.53  \\
UV + BB + S             &      0.66 & 0.61  &  0.74  & 0.97  & 0.60 \\
AV                      &      0.64 & 0.61   &  0.73  & 0.95  & 0.59\\
AV + S                  &     0.73  & 0.76 &  0.80 & 0.96  & 0.69 \\
AV + BB                 &     0.63  & 0.59   &  0.72  & 0.97  & 0.57 \\
AV + BB + S             &     0.68  & 0.64    &  0.77  & 0.97  & 0.63 \\
\hline
\end{tabular}
\label{tab:ablation}
\end{table}
The base configuration using unannotated video (UV) alone achieves a strong F1 score of 0.74, demonstrating that Gemini 2.5 Pro can efficiently extract temporal features from raw visual input. When vehicle speed metadata (S) is added to the UV input, the model shows notable performance gains. Accuracy increases from 0.653 to 0.707, and AUC improves from 0.621 to 0.743. These gains highlight the importance of motion context in improving the model’s visual reasoning. Interestingly, combining unannotated video (UV) with bounding-box coordinates (BB) does not lead to significant performance gains. It lightly reduces both precision and recall. This result suggests that, in the absence of visual annotation overlays, raw coordinate inputs may be less interpretable within the video context. Conversely, configurations using annotated video (AV) consistently outperform UV-based setups. The AV-only baseline achieves 0.671 accuracy, and performance steadily improves as more modalities are added. The best performance is achieved with the AV + S configuration, reaching the highest accuracy of 0.73, an F1 score of 0.80. It also delivers strong precision and AUC values of 0.96 and 0.76, respectively.
These results confirm the importance of structured visual guidance (annotations), ego-vehicle context, and spatial grounding through bounding boxes. The observed incremental improvements in prediction accuracy validate the multimodal design of the BF-PIP model.
\section{Conclusion}
The work presents BF-PIP, a zero-shot framework for predicting pedestrian crossing intent using Gemini 2.5 Pro’s multimodal reasoning. The task is framed as a binary classification problem on short video clips. These clips are either annotated or unannotated and are combined with ego-vehicle speed. Structured prompts provide scene context and behavioral cues to guide the model’s understanding. BF-PIP outperforms existing methods on the JAAD dataset. It performs well in both annotated and raw video settings and generalizes across different traffic scenes. Qualitative and ablation results highlight the value of bounding boxes, motion metadata, and temporal continuity. These results show that large multimodal models, when paired with thoughtful prompt design, can enable reliable intent prediction in autonomous driving systems.

\end{document}